\documentclass[10pt]{article}
\usepackage{fullpage}

\usepackage[numbers]{natbib}

\usepackage{times,amsmath,amsfonts,amssymb,url,color,graphicx}
\usepackage{accents}
\usepackage{wrapfig}
\usepackage{subcaption} 
\usepackage{boxedminipage}

\usepackage{algorithm}
\usepackage{algorithmic}
\algsetup{indent=2em}

\usepackage{etaremune}

\usepackage{relsize}

\usepackage{tikz}
\usetikzlibrary{arrows,decorations.pathmorphing,backgrounds,fit}
\usetikzlibrary{positioning} 
\usetikzlibrary{calc}
\usetikzlibrary{decorations.markings}
\usetikzlibrary{shapes}

\pgfdeclarelayer{l1}
\pgfdeclarelayer{l2}
\pgfsetlayers{l1,l2,main}
\makeatletter
\pgfkeys{%
  /tikz/node on layer/.code={
    \gdef\node@@on@layer{%
      \setbox\tikz@tempbox=\hbox\bgroup\pgfonlayer{#1}\unhbox\tikz@tempbox\endpgfonlayer\egroup}
    \aftergroup\node@on@layer
  },
  /tikz/end node on layer/.code={
    \endpgfonlayer\endgroup\endgroup
  }
}

\def\node@on@layer{\aftergroup\node@@on@layer}
\makeatother

\newtheorem{definition}{Definition}

\newcommand{\BlackBox}{\rule{1.5ex}{1.5ex}}  % end of proof

\newcommand{\reals}{\mathbb{R}}

\newlength{\dhatheight}

\newcommand{\secref}[1]{Section~\ref{#1}}

{
\begin{center}
\begin{boxedminipage}{0.8\linewidth}
\begin{center}
\textbf{\texttt{#1}}
\end{center}
\rm
\begin{tabbing}
....\=...\=...\=...\=...\=  \+ \kill
} %
{\end{tabbing} 
\end{boxedminipage} \end{center} %\vspace{0.3cm}
}

\tikzset{
  lvl1/.style={draw,fill=blue!50,rounded corners=0.5cm,inner sep=5pt,node on layer=l1},
  lvl2/.style={draw,fill=blue!25,rounded corners=0.5cm,inner sep=5pt,node on layer=l2},
  lvl3/.style={draw=blue,fill=white,dashed,rounded corners=0.25cm,align=flush center,text width=10em,inner sep=4pt,minimum height=1cm},
  title/.style={node font=\LARGE,color=white},
  myarrow/.style={latex-latex,ultra thick,blue!80},
}

\title{Vision Zero: Can Roadway Accidents be Eliminated without Compromising Traffic Throughput?}
\author{Shai Shalev-Shwartz, Shaked Shammah, Amnon Shashua}
\date{Mobileye, 2018}

\begin{document}

\maketitle

\begin{abstract}
  We propose a new economical, viable, approach to challenge almost all car
  accidents. Our method relies on a mathematical model of safety and
  can be applied to all modern cars at a mild cost.
\end{abstract}

\section{Introduction}

In 1997 the Swedish Parliament introduced a ``Vision Zero'' policy
that requires reducing fatalities and serious injuries to zero by
2020. 
One approach to reduce the number of serious car accidents, which has
been advocated by the ``Vision Zero'' initiative, is to enlarge the
tolerance to human mistakes by combining regulative and infrastructure
changes. For example, installing speed bumps in urban areas, which
reduces the common speed from 50 kph to
30 kph, may make the difference between a mild injury and a fatality
when a car hits a pedestrian. Another example is not allowing a green
light for two routes at the same time (like ``turn right on red''
scenarios).  The disadvantage of this approach is that it compromises the
throughput of the road system --- for example, reducing the speed
limit from 50 kph to 30 kph increases traveling time by 66\%.

Another approach to reduce the number of car accidents is to
rely on Advanced Driving Assistant Systems (ADAS). For example, a
Forward Collision Warning (FCW) system alerts the driver when the car
is dangerously close to a frontal car and an Automatic Emergency
Braking (AEB) system applies a strong autonomous braking at the last
moment in case an accident is likely to happen. A recent study of the
Insurance Institute for Highway Safety (IIHS) shows that vehicles
equipped with FCW and AEB systems have a 64\% fewer front-to-rear
crashes with injuries~\cite{IIHS}. The advantage of the ADAS approach
is that it does not affect the throughput of the road system.

The main goal of this paper is to propose a roadmap for reaching an
ADAS system that can substantially reduce fatalities and serious
injuries, at a reasonable cost, while sustaining the usefulness and
throughput of the road system. We emphasize that the industry as a
whole keeps enhancing ADAS systems and comes up with novel new
approaches (a notable example is Toyota's Guardian
\cite{guardian}). Our goal is to put on the table a complete proposal
with a clearly stated goal, which is fully accessible to the public
and to regulatory bodies, and which comes with some formal mathematical guarantees.

Our starting point is exiting AEB systems. As mentioned previously,
AEB systems already eliminate roughly 64\% of front-to-rear
crashes. While this is an impressive achievement, we believe that the
main reason some crashes are not eliminated is because AEB is an
``emergency'' system rather than a ``preventive'' system. That is, AEB
systems are designed to react to an imminent collision (hence the term
``emergency'' in the acronym) rather than to prevent a dangerous
situation to occur in the first place.  Indeed, the AEB test
specifications by regulatory bodies (like ENCAP \cite{encap}) are
geared towards avoiding or mitigating a collision that is expected if
no action is taken. To balance safety and comfort, normally the brake
activation is applied at a very short time-to-collission (TTC) in order
to reduce the chance of a false detection causing an unexpected brake
activation\footnote{Car makers adopt a variety of "braking profiles", but as shown in \cite{ADAC} all braking profiles begin braking at TTCs not higher than 2s.}. As a result, AEB avoids collisions at relatively low
speeds and somewhat mitigates collisions at higher speeds but even
those are much lower than highway driving speeds. Electronic Stability
Control (ESC), on the other hand, is designed to reduce skidding,
operates in a ``preventive'' fashion by limiting the maneuverability
of the vehicle. Most drivers do not feel constrained and mostly are
not even aware that ESC was activated while those who seek full
unencumbered control of the vehicle can always deactivate ESC if they
so desire.

The first step in our roadmap is therefore a ``preventive'' collision
avoidance system in which a mild brake activation is applied before
the imminent danger occurs while keeping the comfort of driving
intact. We wish to do so while providing \emph{formal guarantees} of
safety in the sense of proving that if all cars comply with the
proposed system then collisions rates will drop considerably below
current AEB rates. Our approach embraces the ``prevention'' rational
of the ``Vision Zero'' initiative --- unlike AEB systems, that are
activated at the last moment, we would like to keep vehicles from
entering dangerous situations in the first place. A notable advantage
of this ``preventive'' approach over AEB is that unlike AEB systems
that prevent an accident by braking very strongly (1.5g in some
systems) and as a result might cause someone else to hit us from
behind, the preventive approach will eliminate the danger, of being hit from behind,  in advance
by applying a mild brake. At the same time, we embrace the ``keep the
normal flow of traffic'' rational approach of ADAS systems in general,
and AEB in particular, as we refrain from reducing speed without
distinction. Instead, we intervene in the driver's decisions only when
he/she is not ``careful'' according to the specific scenario.

At the heart of our approach is the Responsibility Sensitive Safety
(RSS) model described in \cite{RSS}. RSS formalizes an interpretation
of reasonable human common sense. RSS is a rigorous mathematical model
formalizing an interpretation of a set of principles of reason, and has been designed to
achieve several goals: first, the interpretation of how humans interpret {\it caution}. Second, the interpretation should lead to a desired result (we call
``Utopia'') that, if all agents comply with RSS parameters, 
accidents, resulting from decision making processes, could be eliminated. Third, the interpretation should be
\emph{useful}, meaning it will enable  agile driving rather than
enforcing an overly-defensive driving which inevitably would block
traffic and in turn limit the scalability of system
deployment. Finally, the interpretation should be \emph{efficiently verifiable} in the sense that we can rigorously prove that a car will
always obey the interpretation of human caution without the need to role out future actions of all agents involved. 

Originally, RSS has been designed as a safety seal for the decision
making process of self driving cars. However, we show in this paper
how a variant of RSS can be leveraged to ADAS, to enhance the safety of human drivers as
well while providing 
\emph{formal mathematical guarantees} for safety when all agents comply with RSS. In a nutshell, RSS introduces a parameter-based methodology for defining safe distances which when breached constitute a ``dangerous'' situation. Safe distance was originally couched in a specific "braking profile" that fits a robotic control. In this paper we extend the definitions of RSS to include any braking profile and in particular we propose a jerk-bounded profile that allows for relatively large TTC yet supports smooth braking. The benefit of couching a braking profile within the RSS framework is that all the formal guarantees associated with RSS apply as well. Thus, rather than proposing {\it some heuristic\/} of a braking profile, RSS enables a preventive braking system with formal guarantees. Furthermore, we argue that
the proposed RSS-for-ADAS system may be similar to ESC in the sense where most
drivers would have no problem with mild interventions of the car as a
safety mechanism, and the few drivers which would like
full unabated control for competitive driving will be able to shut off the system and be exposed to a higher level of accident risk. 

The rest of the sections describe the roadmap for an ADAS system that
enforces compliance with RSS rules. We envision two stages of
deployment where at first we propose merely an enhancement of existing
AEB systems in the sense of handling front-to-rear crashes only using
existing front-facing sensing (camera or camera+radar). Our proposal
for stage 1, can lead to a significantly higher elimination rate of
front-to-rear crashes, comparing to existing AEB systems. The main
improvement of stage 1 is due to the preventive approach, which will
allow technology providers to balance the false positive / false
negative tradeoff in a better way. By considering the sensing
capabilities of existing systems, we estimate that the elimination
rate of front-to-rear accidents will be roughly $99\%$. For the second stage we envision a
full surround camera sensing fused with a crowd sourced map, to enable
the full implementation of RSS. The advantage of the full
implementation is that RSS comes with the elegant mathematical guarantee,
stating that if all players fully comply with RSS rules, then accidents resulting from driving decision making process 
would become rare, thus achieving Vision Zero~\footnote{As we explain later,
  accidents can still occur if some agents do not comply with RSS
  rules due to hardware failure, sensing error, software bugs or some ``act of
  god''. But, it is reasonable to estimate that such failure will
  happen in less than $1\%$ of the dangerous situations, and under
  this assumption, at least $99\%$ of the accidents would be
  eliminated. } for all possible crashes. The cost of installing a
camera surround system, in high volume, would be around $5-10$ times
the cost of existing AEB systems which is negligible compared to the
cost of accidents to society.

\section{Stage I: preventing front-to-rear crashes by Automatic
  Preventive Braking (APB)}

The goal of stage I is to prevent front-to-rear crashes as an
enhancement of existing AEB systems. As of 2018, AEB systems are
activated when the Time-To-Collission (TTC) to the front vehicle/pedestrian
is very small. Typically, a number between $1$ to $2$ seconds. The
time to contact is the distance between the cars divided by their
relative speed. It follows that if the front and rear vehicles are
driving at the same speed, the TTC is infinity. To illustrate why this
approach is problematic, consider an extreme case, in which the rear
vehicle is driving slightly faster than the front vehicle. Say, the
rear vehicle is driving at $10\,m/s$ ($=36$ kmh) while the front
vehicle is driving at $9.99\,m/s$. If the distance between them is
$x$, then the TTC is $x/0.01 = 100\,x$. Taking an AEB system which is
being activated at TTC of $2$ seconds, we obtain that the system will
not be activated as long as the distance is larger than $2/100 m = 2
\,cm$. It follows that driving behind a car at a speed of $36$ kmh and at a
distance of $3$ cm is considered ``safe'' by an AEB system. This
does not feel right.

The source of the aforementioned problem in AEB systems is that they
do not consider a ``what would happen if'' type of reasoning. Indeed,
if the front car would suddenly brake, the AEB system would need a
time to respond to the change and it will most likely be too late. The
common sense of a safe human driver is to be a little bit paranoid ---
a good driver keeps a safe distance from a frontal car so as to be
ready for the unexpected. Of course, being over protective is also not 
good, as it leads to an extremely defensive driving. The secret sauce
is to be ready for the unexpected, yet reasonable, events, while ignoring
completely un-reasonable events. 

RSS~\cite{RSS} is a mathematical, interpretable, model, formalizing the ``common
  sense'' arguments above. In particular, RSS formalizes
\begin{itemize}
\item What is a \emph{dangerous} situation ?
\item What is the \emph{proper response} to a dangerous situation ?
\item What are the \emph{reasonable assumption} a good driver can make
  on the future behavior of other road agents ?
\item What does it mean to be \emph{cautious} ?
\end{itemize} 
For a detailed exposition of RSS we refer the reader to \cite{RSS}. We
do not intend to repeat all the definitions in this paper, but only to
introduce a generalization of RSS that will be useful for human drivers.

We first consider the simplest case of a front car, $c_f$,
driving in front of a rear car, $c_r$, where both cars driving at the
same direction along a straight road\footnote{see \cite{RSS} on homomorphism from general to straight roads.}, without performing any lateral
maneuvers\footnote{to support lateral maneuvers see Sec.\ref{sec:stage2}.}.  RSS defines the notion of a dangerous situation by
relying on the following definition:
\begin{definition}[RSS Safe Distance] \label{def:dmin} 
A \emph{longitudinal distance} between a
  car $c_r$ that drives behind another car $c_f$, where both cars are
  driving at the same direction, is \emph{safe} w.r.t. a response time
  $\rho$ if for any braking of at most $a_{\max,\mathrm{brake}}$,
  performed by $c_f$, if $c_r$ will accelerate by at most
  $a_{\max,\mathrm{accel}}$ during the response time, and from there
  on will brake by at least $a_{\min,\mathrm{brake}}$ until a full
  stop then it won't collide with $c_f$.
\end{definition}

Relying on the above definition, RSS states that:
\begin{itemize}
\item A situation is \emph{dangerous} if the distance is not safe.
\item The \emph{proper response} to a dangerous situation for the rear
  car is to brake (after a response time) by at least
  $a_{\min,\mathrm{brake}}$ until either reaching a full stop or
  gaining again a safe distance
\end{itemize} 

Crucially, RSS's proper response does not depend on the underlying
driving policy. In fact, it can be embedded on top of any driving
policy. Our main observation is that we can embed RSS on top of a
human driving policy. Basically, whenever a human driver brings the
car to a non-safe distance, the RSS-ADAS system will apply braking so
as to bring the car back to a safe distance.

The problem with this naive implementation is that applying a strong
brake without any warning might be dangerous by itself.\footnote{This
  is one of the main reason why existing AEB systems are tuned
  to have an extremely low probability of false positives. The price
  for extremely low false positive rate is a much higher false
  negative rate (meaning, the system fails to detect a car which is at
  a non-safe distance).}  To tackle this problem, we first propose a
generalization of RSS (\secref{sec:genRSS}), and then we specify a
particular member of this RSS family that enables to intervene in
advance but in a smooth manner without inconveniencing the human driver
(\secref{sec:jbCRSS}). Another advantage of this approach is that it
mitigates the danger of false positives.

\subsection{Generalized RSS} \label{sec:genRSS}

In the original definition of RSS, the rear car is assumed to
accelerate during the response time and then to brake until reaching a
full stop. This is an example of a \emph{braking profile}. More
generally:
\begin{definition}[Braking profile]
A braking profile, $B$, is a mapping from initial kinematic state of
the car (mainly, initial velocity, $v_0$, and acceleration, $a_0$) to
a pair $T_b, v$, s.t. $v : [0,\infty) \to \reals$ is the future
velocity of the car and $T_b < 0$ is the first time in which $v(t)=0$. 
\end{definition}

Some examples for braking profiles are given below:
\begin{itemize}
\item The braking profile applied by the front car in the
original definition of RSS is defined by $T_b =
\frac{v_0}{a_{\max,\mathrm{brake}}}$ and $
v(t) = \max\{v_0 - t
a_{\max,\mathrm{brake}}, 0\}$. 
\item The braking profile applied by the rear car in the
original definition of RSS is defined by $T_b = 
\frac{v_0 + \rho a_{\max,\mathrm{accel}}}{a_{\min,\mathrm{brake}}}$,
and \[
v(t) = \begin{cases}
v_0 + t a_{\max,\mathrm{accel}} & \mathrm{if}~ t \le \rho \\
v_0 + \rho a_{\max,\mathrm{accel}} - (t-\rho)\,a_{\min,\mathrm{brake}}
& \mathrm{if}~ t \in (\rho,T_b) \\
0 & \mathrm{otherwise}
\end{cases}
\]
\end{itemize}

We now define a generalization of RSS's safe distance.
\begin{definition}[Generalized RSS Safe Distance w.r.t. Braking
  Profiles $B_f,B_r$] \label{def:dmin_gen} 
A \emph{longitudinal distance} between a
  car $c_r$ that drives behind another car $c_f$, where both cars are
  driving at the same direction, is \emph{safe} w.r.t. braking
  profiles $B_f,B_r$ if in case the front car applies braking profile
  $B_f$ and the rear car applies braking profile $B_r$, then the cars
  will reach a full stop without colliding. 
\end{definition}

The \emph{proper response} is defined as follows:
\begin{definition}[Proper response] \label{def:pr_gen} 
Suppose the first time that the distance between $c_f$ and $c_r$ is
non-safe is $t_0$, and w.l.o.g. set $t_0 = 0$. Then, the proper
response for the front car is to have its velocity at least $v_f(t)$,
where $v_f$ is the velocity defined by $B_f$, and the proper response
for the rear car is to have its velocity at most $v_r(t)$, where $v_r$
is the velocity defined by $B_r$. 
\end{definition}

It is easy to verify that if both cars apply proper response then
there will be no accident. 

\subsection{Jerk-bounded Braking Profile} \label{sec:jbCRSS}

In this section we describe the jerk-and-acceleration-bounded braking
profile, denoted $B_j$. The idea is that we start decreasing our
acceleration linearly (with slope $j_{\max}$), until reaching a max
brake parameter (denoted $a_{\min,\mathrm{brake}}$), and then we
continue to brake with a constant deceleration until reaching zero
velocity. In the following we provide closed form formulas for
calculating this braking profile. 

\paragraph{Braking with a Constant Jerk}

Suppose we start braking at jerk of $j_{\max}$. Suppose also that we can
immediately arrive to zero acceleration (leave the throttle), hence below we
assume $a_0 \le 0$. Then, the dynamics of the car is as
follows:
\begin{align*}
a(t) &= a_0 - j_{\max} t \\
v(t) &= v_0 + \int_{\tau = 0}^t a(\tau) d \tau = v_0 + \left[ a_0 \tau - \frac{1}{2} j_{\max}
       \tau^2 \right]_0^t   = v_0 +  a_0 t - \frac{1}{2} j_{\max} t^2 \\
x(t) &=  x_0 + \int_{\tau = 0}^t v(\tau) d \tau = x_0 + \left[  v_0 \tau + \frac{1}{2} a_0
       \tau^2 - \frac{1}{6} j_{\max} \tau^3\right]_0^t   = x_0 +
       v_0\,t + \frac{1}{2} a_0 t^2 - \frac{1}{6} j_{\max} t^3
\end{align*}

\paragraph{Braking profile}

Based on these equations, braking distance will be defined as
follows. Let $T$ be the first time in which either $a(T) = -a_{\min,\mathrm{brake}}$
or $v(T) = 0$, that is, 
\[
T = \min\{T_1,T_2\} ~~~\textrm{where}~~~ T_1 =
\frac{a_0+a_{\min,\mathrm{brake}}}{j_{\max}} ~~,~~ T_2 = 
  \frac{a_0 + \sqrt{a_0^2 +2 j_{\max} v_0}}{j_{\max}} ~.
\]
The time for reaching full brake is
\[
T_b = \begin{cases}
T_2 & \mathrm{if}~ T = T_2 \\
\frac{v_0 +  a_0 T - \frac{1}{2} j_{\max}
  T^2}{a_{\min,\mathrm{brake}}} & \mathrm{otherwise}
\end{cases}
\]
And the speed is as follows:
\[
v(t) = \begin{cases}
 v_0 +  a_0 t - \frac{1}{2} j_{\max} t^2 & \mathrm{if}~ t \le T \\
v_0 +  a_0 T - \frac{1}{2} j_{\max}
  T^2 - (t-T) a_{\min,\mathrm{brake}} & \mathrm{if}~ t \in (T,T_b) \\
0 & \mathrm{otherwise}
\end{cases}
\]

\paragraph{Braking distance}
Finally, the car will move the following distance until reaching a full stop:
\[
\left[v_0\,T + \frac{1}{2} a_0 T^2 - \frac{1}{6} j_{\max} T^3 \right]
+ \frac{\left(v_0 +  a_0 T - \frac{1}{2} j_{\max}
    T^2\right)^2}{2 a_{\min,\mathrm{brake}}}
\]

\subsection{Safe distance w.r.t. $B_f$ and $B_j$}

Suppose that the front car brakes with braking profile $B_f$ (braking
with a constant deceleration of $a_{\max,\mathrm{brake}}$), and the
rear car relies on braking profile $B_j$. Then, the safe distance
formula is 
\[
  \left[ \left[v_0\,T + \frac{1}{2} a_0 T^2 - \frac{1}{6} j_{\max} T^3 \right] + \frac{\left(v_0 + a_0 T - \frac{1}{2} j_{\max}
    T^2\right)^2}{2|a_{\min,\mathrm{brake}}|} - \frac{v_f^2}{2|a_{\max,\mathrm{brake}}|}
\right]_+
\]

\subsection{The Automatic Preventive Braking (APB) System}

As soon as the distance becomes non-safe (according to the jerk-based
formula), the APB system will start to brake
with a jerk of $j_{\max}$ until one of the following happens: (1) the
car stops (2) the distance becomes safe, or (3) the driver turns off
the RSS system (by a button or some other mechanism).

An inductive proof, similar to the one given in \cite{RSS}, can show
that if we start with a safe distance from a front vehicle, and apply
the proper response of the APB System, then under the
assumption that the front vehicle will not brake stronger than
$a_{\max,\mathrm{brake}}$, we will never hit the front vehicle from behind.

\subsection{A better balance between false positives
  and false negatives}

It is worth noting that the same hardware that implements AEB nowadays
can be adjusted to implement APB. Since AEB systems brake abruptly at
the last seconds before a crash, an activation of the AEB system is
scary and dangerous (someone might hit us from behind), and hence AEB
systems are tuned to have an extremely small rate of false positives
(a false positive event is when the system detects that we are going
to crash into a ``ghost'' vehicle, which is not really there). The
price for an extremely small rate of false positives is a higher rate
of false negatives (a false negative event is when the system fails to
detect a car, even though we are going to crash into it). Furthermore,
since false positive events are much more dangerous when driving at a
high speed, AEB systems are only activated when driving at a low
speed. 

In contrast, because the braking profile of APB is jerk bounded, the
danger of false positives is much milder (the driver will not
be jolted, and a car from behind will also not be surprised
because of the bound on the jerk). Therefore, we can tune the system
to have much less false negatives than an AEB system, and we can also
activate the APB system when driving at a high speed.

%\section{Stage II: reducing unsafe cut-ins using Lane Keeping Assist}
%
%In the previous section we have described Stage I of the system, which
%deals with front-to-rear collisions when a rear car continuously
%follows a frontal car. However, it doesn't prevent car from performing
%a reckless cut-in. In many cases, a reckless cut-in will be due to a
%driver that wasn't paying attention. Suppose that we do not allow the
%car to move laterally in the lane unless the driver switch on the turn
%signal. This can be achieved by existing Lane Keeping Assist
%systems. As a result, we will eliminate all types of reckless cut-in
%that follow from a non-attentive driver. Of course, we will not
%eliminate accidents that follows from a driver that turns
%on the turn signal and actively performs a reckless cut-in.
%
%\section{Stage III: preventing unsafe cut-in using surround sensing}
%
%In Stage III of the system we can enhance the Lane Keeping Assist
%system and do not allow a driver to perform a reckless cut-in, even if
%it turns on the signaling light.  This requires a surround sensing system.

\section{Stage II: Beyond Front-Rear Accident Prevention}\label{sec:stage2}

In stage II of the system, we require a surround (camera-based due to
the required resolution) sensing system and a crowd sourced High
Definition (HD) map. The surround system enables to prevent a car from
performing a reckless cut-in and to merge in junctions unsafely.
The major contribution of the map is that the system has full
knowledge on where to expect potential dangers: it knows in advance
where there should be lanes that might merge to our lane, who has the
priority, where to expect traffic lights, where to expect occluded
pedestrians, etc.  Planning in advance enables the system to adjust
speed mildly in advance and not to be surprised. In addition, a map
enables to know the geometry of the road even without explicitly
detecting lane marks. Finally, RSS is a formalization of
reasonable common sense in such scenarios. With the use of  a crowd sourced
map, we can gain access to road conditions and the actual data on the behavior of road users at every
road, thus helping regulators to define the appropriate rules for every
road (for example, to specify ``school zone'' areas in an automatic
way). 

Unlike geo-fenced autonomous driving applications limited to particular areas, an ADAS
system would require the support of an HD-map in all driving locations.  An
 HD map that can offer coverage of the entire globe and is perpetually maintained fresh and updated by means of crowd sourced data. 
 With a surround sensing and a map, we can implement the full stack of
RSS. The details can be readily inferred following \cite{RSS} and the exposition of Stage-I above.

\section{Utopia is (almost) Possible --- Toward Reaching Vision Zero}

The premise of RSS is that ``utopia is possible'', in the sense that
if all agents fully comply with RSS's proper response,\footnote{By ``following
  proper response'' we mean that the agents' actual decelerations match
  the requirements of proper response.}
 then accidents resulting from wrongful driving decisions would become rare. And so, geographic regions enforcing  RSS on all vehicles
(by regulation), will benefit from a substantially high level of road safety. It should be noted that accidents can still occur as a result of conditions outside the driving decision making process, such as  accidents 
caused by vehicles that do not apply proper response due to a hardware
failure (e.g. malfunctioning braking system) or due to perception mistakes (e.g., a system does not detect
a car at a non-safe distance in front of us).  However, suppose that
the probability of such a failure is one in every 100 dangerous
situations. While this level of accuracy is insufficient for
autonomous driving (an accident once in every 100 dangerous situation
is a very bad result), it can be shown that if this failure rate is maintained in ADAS systems, where the driver is still active and responsible, we can obtain an elimination rate of roughly 99\% of car accidents. This is a huge step toward
fulfilling the ``Vision Zero'' initiative.

\bibliographystyle{plain}
%\bibliography{bib}

\end{document}